\documentclass[lettersize,journal]{IEEEtran}
\usepackage[T1]{fontenc}

\usepackage{amsmath,amsfonts}
\usepackage{algorithmic}
\usepackage{algorithm}
\usepackage{array}
\usepackage{subcaption}
\usepackage{textcomp}
\usepackage{stfloats}
\usepackage{url}
\usepackage{verbatim}
\usepackage{graphicx}
\usepackage{cite}
\usepackage{xcolor}
\usepackage[numbers,sort&compress]{natbib}
\usepackage{booktabs}
\usepackage{multirow}
\usepackage{graphicx}

\begin{document}


\title{Real-World Deployment of Cloud-based Autonomous Mobility Systems for Outdoor and Indoor Environments} 

\author{ Yufeng~Yang$^{1}$\textsuperscript{\textdagger},
         Minghao~Ning$^{1*}$\textsuperscript{\textdagger},
         Keqi~Shu$^{1}$\textsuperscript{\textdagger},
         Aladdin~Saleh$^{2}$,
         Ehsan~Hashemi$^{3}$,~\IEEEmembership{Senior Member,~IEEE},~and
         Amir~Khajepour$^{1}$,~\IEEEmembership{Senior Member,~IEEE}

\thanks{\textsuperscript{\textdagger} Indicates equal contribution.}
\thanks{* Corresponding author}
\thanks{Y. Yang, M. Ning, K. Shu and  A. Khajepour are with the Department of Mechanical and Mechatronics Engineering, University of Waterloo, Ontario, N2L 3G1, Canada (e-mail: \{f248yang, minghao.ning, keqi.shu, a.khajepour\}@uwaterloo.ca)}
\thanks{A. Saleh is with the Technology Partnerships and Innovations, Rogers Communications, Canada Inc., Toronto, Ontario, M4Y 2Y5, Canada (email: aladdin.saleh@rci.rogers.com)}
\thanks{E. Hashemi is with the Mechanical Engineering Department, University of Alberta, Alberta, T6G1H9, Canada (e-mail: ehashemi@ualberta.ca)}
}

\markboth{Journal of \LaTeX\ Class Files,~Vol.~14, No.~8, August~2021}%
{Shell \MakeLowercase{\textit{et al.}}: A Sample Article Using IEEEtran.cls for IEEE Journals}


\maketitle

\begin{abstract}

Autonomous mobility systems increasingly operate in dense and dynamic environments where perception occlusions, limited sensing coverage, and multi-agent interactions pose major challenges. While onboard sensors provide essential local perception, they often struggle to maintain reliable situational awareness in crowded urban or indoor settings. This article presents the Cloud-based Autonomous Mobility (CAM) framework, a generalized architecture that integrates infrastructure-based intelligent sensing with cloud-level coordination to enhance autonomous operations. The system deploys distributed Intelligent Sensor Nodes (ISNs) equipped with cameras, LiDAR, and edge computing to perform multi-modal perception and transmit structured information to a cloud platform via high-speed wireless communication. The cloud aggregates observations from multiple nodes to generate a global scene representation for other autonomous modules, such as decision making, motion planning, etc. Real-world deployments in an urban roundabout and a hospital-like indoor environment demonstrate improved perception robustness, safety, and coordination for future intelligent mobility systems.

\end{abstract}


\section{Introduction}




As cities grow denser and operating environments become increasingly complex, autonomous robots are expected to operate not in isolation, but as integral components of dynamic mobility ecosystems. From urban streets shared with pedestrians and vehicles \cite{pechinger2024threshold,mo2024enhanced} to public building corridors crowded with people \cite{karthik2024self,sahoo2024autonomous}, robots must navigate areas that are inherently occlusion-prone, uncertain, and socially interactive. Despite rapid advances in onboard perception, today’s autonomous systems remain fundamentally constrained by a self-contained sensing structure: they perceive the world only from their own limited viewpoint.

This self-reliant architecture creates a structural bottleneck. The use of onboard cameras and LiDAR may cause restricted fields of view, frequent occlusions, and geometric blind spots. In dense urban traffic, a vehicle cannot see beyond large obstacles or around corners. In indoor facilities such as hospitals and warehouses, narrow corridors and dynamic human flows may further degrade perception reliability. These limitations compound, making coordination and system-level safety increasingly difficult to guarantee.


In this article, we introduce a generalized cloud-based architecture designed for both outdoor \cite{ning2025coinfra} and indoor environments \cite{ning2025sapcopesocialawareplanningusing}, called Cloud-based Autonomous Mobility (CAM). The proposed framework deploys distributed Infrastructure Sensor Nodes (ISNs) equipped with multi-modal perception and edge computing capabilities. These nodes provide high-mounted, wide-field environmental awareness that complements onboard sensing, reduces occlusions, and enhances temporal consistency. Unlike traditional infrastructure sensing systems that primarily serve traffic monitoring or offline data collection \cite{pechinger2023roadside,zhang2025mic,zimmer2023infradet3d,zimmer2024tumtraf,dasgupta2024object,jeong2024gdtm}, the proposed framework is designed for operational integration. It directly supports autonomous navigation by supplying real-time object detection, tracking, and contextual understanding to other modules such as decision-making, motion planning, etc.

\begin{figure}
    \centering
    \includegraphics[width=0.49 \textwidth]{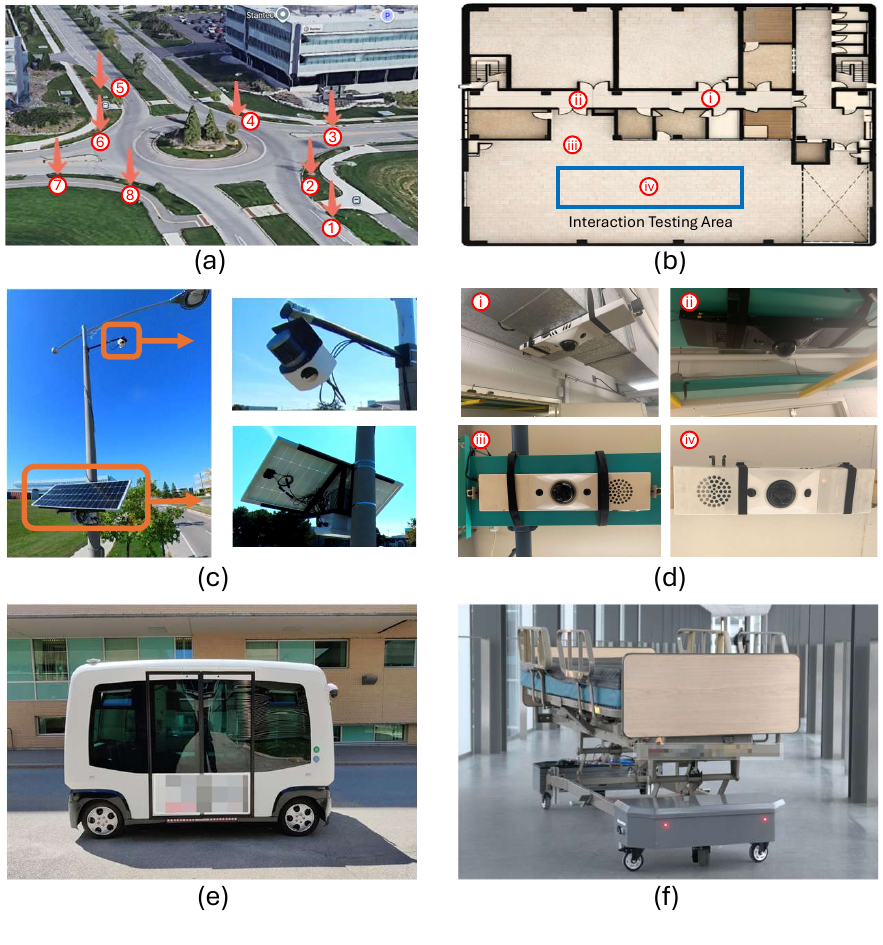}
    \caption{Infrastructure Sensor Node (ISN) deployment layout. (a) Outdoor environment with eight ISNs deployed around a roundabout. (b) Indoor environment with four ISNs covering part of a laboratory area and an adjacent hallway. (c) Individual outdoor ISN unit. (d) Individual indoor ISN unit. (e) Autonomous shuttle bus (f) Indoor autonomous mobile robot}
    \label{Fig: framework}
\end{figure}

The proposed CAM framework has the following advantages:

\begin{itemize}

    \item \textbf{Shared environmental perception for complex mobility environments:}
    By elevating sensing capabilities to infrastructure nodes, the proposed system provides wide-area perception that complements and augments onboard sensing. High-mounted ISNs reduce occlusions and geometric blind spots while maintaining continuous observation of dynamic agents. This shared perception layer improves situational awareness and perception robustness in crowded urban spaces as well as confined indoor environments.

    \item \textbf{Adaptive accuracy–latency trade-offs:}
    The CAM framework integrates edge computing and high-speed communication (e.g., 5G) to support real-time cooperative perception with predictable latency. Its modular perception pipeline allows flexible fusion strategies that balance accuracy, bandwidth, and computational delay.  

    \item \textbf{Generalized cloud-based framework:}
    We introduce a generalized cloud-enabled sensing framework that enables collaboration between intelligent infrastructure and autonomous robots. The framework supports both outdoor and indoor environments, enabling scalable multi-node sensing through synchronized observations, delay-aware fusion, and global object tracking. The generated cooperative perception can be used for any other modules, such as decision-making, motion planning, control, etc.

\end{itemize}

The structure of this article is organized as follows. The overview of the proposed CAM framework and hardware design of ISNs is provided in the next section. Section "Demonstrated Advantages of the Infrastructure-based Sensing" provides a detailed explanation and supporting results to demonstrate the advantages of using the infrastructure-based sensing structure. Application case studies for both outdoor and indoor using the proposed CAM framework are provided in "Case Study: Outdoor application" and "Case Study: Indoor application", respectively. Finally, the conclusion of the article can be found in the last section.

\section{System Architecture and Hardware Design}  \label{Chap: Framework Overview}


\begin{figure*}
    \centering
    \includegraphics[width=0.9 \textwidth]{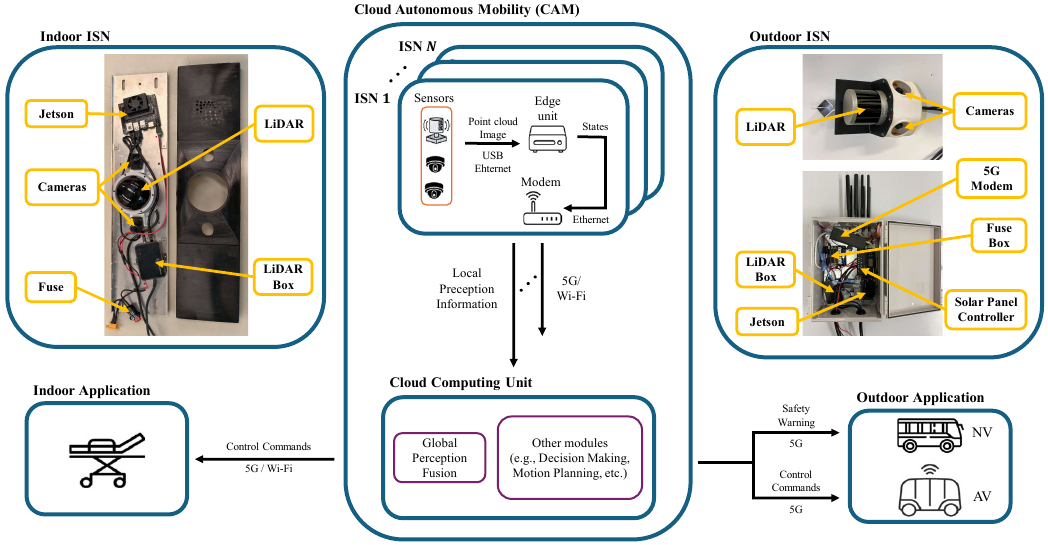}
    \caption{Information flow of the overall framework.}
    \label{Fig: Info_Flow}
\end{figure*}

\subsection{System-Level Information Flow}

Fig. \ref{Fig: Info_Flow} illustrates the overall CAM framework and the hardware configurations of the indoor and outdoor ISNs. The system follows a distributed perception architecture in which multiple ISNs perform local multi-modal sensing and edge processing, and transmit structured perception outputs to a cloud computation for global fusion and other modules.

Specifically, each ISN integrates cameras and LiDAR to capture visual and geometric information. Raw sensor data are processed locally on an embedded edge computing platform (e.g., NVIDIA Jetson), where multi-modal fusion, object detection, and tracking are executed in real time. Instead of streaming raw data, the node transmits compact semantic representations via Wi-Fi or 5G.

A cloud computing unit then collects information from multiple ISNs to construct a global perception result. This fused perception module supports other autonomous modules, such as decision making, motion planning, and control, etc. Finally, the calculated control commands or safety warnings are sent to mobile agents operating in outdoor or indoor environments.

\subsection{Outdoor Intelligent Sensor Node Design}

The outdoor ISNs are designed for unstructured and weather-exposed environments such as roadways or campus streets. It includes:

\begin{itemize}
    \item A 3D LiDAR.
    \item Two cameras.
    \item Edge computing unit (Jetson).
    \item 5G modem for low-latency communication.
    \item Fuse box and solar power controller
    \item Weatherproof enclosure
\end{itemize}

The outdoor ISNs are mounted to roadside lightpoles, and the deployment prioritizes extended sensing range, robustness to weather changes and 5G communication.

\subsection{Indoor Intelligent Sensor Node Design}
The indoor ISN extends the same sensing principle for structured yet occlusion-prone environments such as hospitals and logistics facilities. It consists of:
\begin{itemize}

    \item A 3D LiDAR.
    \item     Two cameras.
    \item Edge computing unit (Jetson).
    \item Power distribution and a cover.
\end{itemize}

The indoor ISNs are mounted to the ceiling to provide elevated, wide-angle coverage in corridors and intersections. Indoor deployment emphasizes stable power supply, short-to-medium sensing range, and reliable Wi-Fi connectivity.

\section{Infrastructure-based Sensing and Advantages} \label{Chap: Advantages}

\begin{figure*}[t]
    \centering
    \begin{subfigure}[b]{0.49\textwidth}
        \centering
        \includegraphics[width=\textwidth]{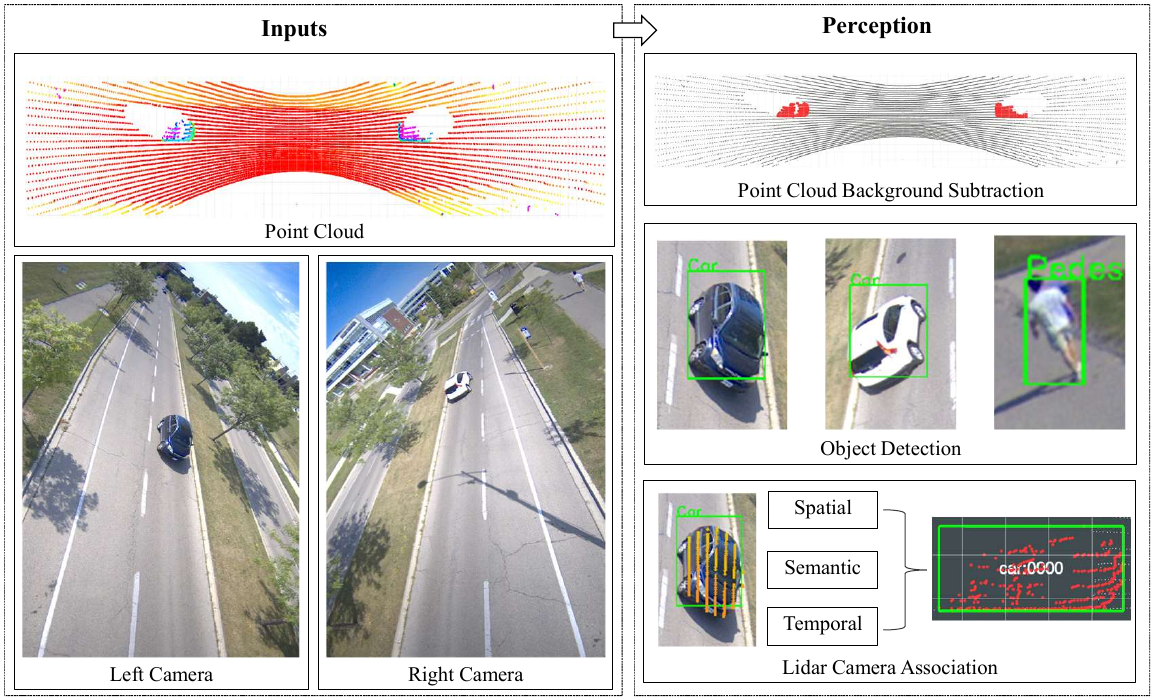}
        \caption{Outdoor Local Perception Structure}
        \label{fig:image1}
    \end{subfigure}
    \hfill
    \begin{subfigure}[b]{0.49\textwidth}
        \centering
        \includegraphics[width=\textwidth]{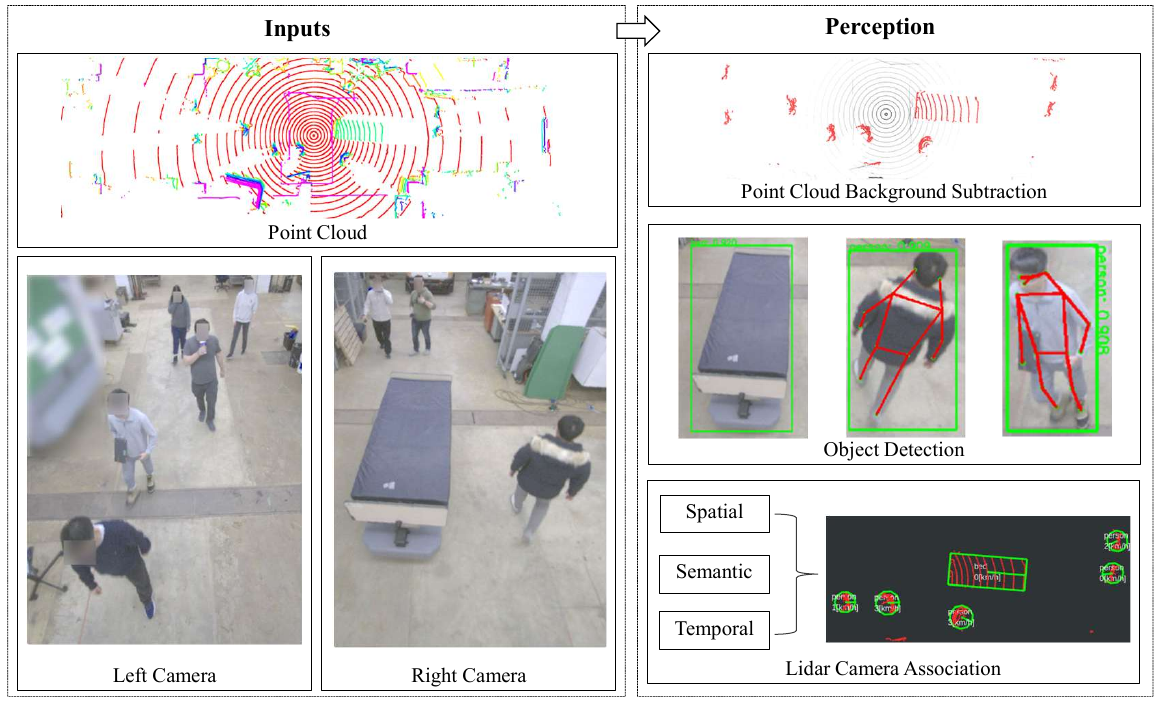}
        \caption{Indoor Local Perception Structure}
        \label{fig:image2}
    \end{subfigure}
    \vspace{0.1cm}

    \begin{subfigure}[b]{0.49\textwidth}
        \centering
        \includegraphics[width=\textwidth]{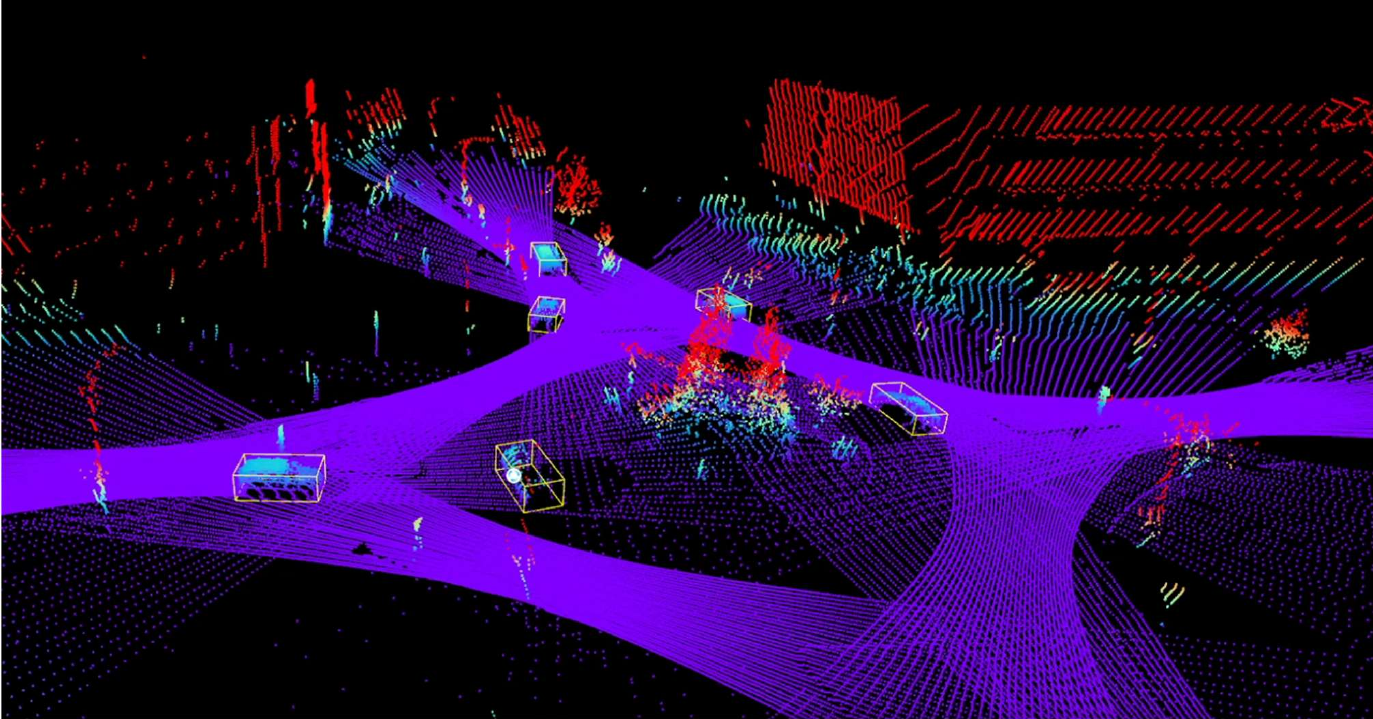}
        \caption{Outdoor Global Perception Visualization}
        \label{fig:image3}
    \end{subfigure}
    \hfill
    \begin{subfigure}[b]{0.49\textwidth}
        \centering
        \includegraphics[width=\textwidth]{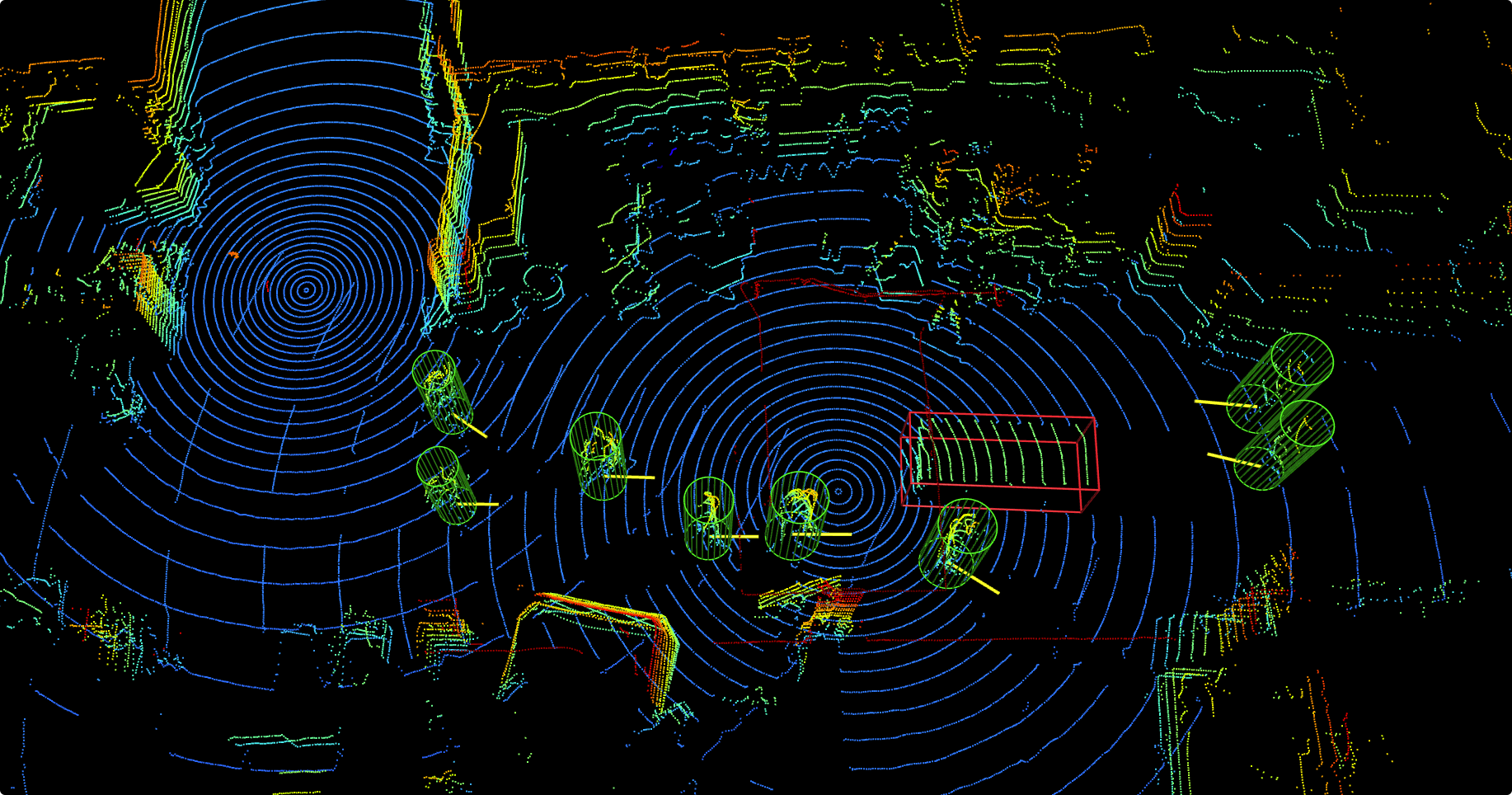}
        \caption{Indoor Global Perception Visualization}
        \label{fig:image4}
    \end{subfigure}

    \caption{Overview of the Perception and Fusion Pipeline.}
    \label{fig:overall_label}
\end{figure*}

\subsection{Overview of the Perception and Fusion Pipeline}
The overall perception and fusion pipeline of the CAM system is shown in Fig.~\ref{fig:overall_label}. Each ISN has its own LiDAR and camera data to generate object detections, which are then transmitted to the cloud for global fusion and tracking~\cite{ning2024enhancingindoormobilityconnected, ning2025coinfra}. 

The local perception will first filter the point cloud by the HD map, and different strategies of LiDAR camera fusion can be used to generate the object detection considering the LiDAR--camera projection in Eqn.~(\ref{eq:cam_proj}): early fusion, which fuses the point cloud and camera image directly as Bird's Eye View (BEV) features and then feeds into a detection network; late fusion, which processes the point cloud and camera image separately to generate two sets of detections and then associate them together; and intermediate fusion, which first generates clusters from the point cloud and then refine them with camera detections.

\begin{equation}
    s \begin{bmatrix} x_p \\ y_p \\ 1 \end{bmatrix} = H_{3\times4}\,\mathbf{p}_w, \qquad H = K[R \mid t],
    \label{eq:cam_proj}
\end{equation}
where $\mathbf{p}_w = [x_w, y_w, z_w, 1]^\top$ denotes a 3D point in homogeneous world coordinates, $(x_p,y_p)$ is the corresponding image-plane coordinate, and $s$ is the projective depth scale introduced by perspective projection. Here, $K \in \mathbb{R}^{3\times3}$ is the camera intrinsic matrix, $R \in \mathbb{R}^{3\times3}$ is the rotation matrix from the world frame to the camera frame, and $t \in \mathbb{R}^{3}$ is the translation vector of the world origin expressed in the camera frame.

For indoor human-centric applications, the pipeline additionally refines per-joint 3D positions via Jacobian-based uncertainty propagation and skeletal bone-length constraints~\cite{ning2025sapcopesocialawareplanningusing}.

At the cloud, all ISN streams are NTP-synchronized. After each node produces object-level 3D detections in a common reference frame, cross-node detections are merged by NMS for rigid objects or by maximum-likelihood pose refinement for pedestrians, and an EKF tracks each object’s planar state $\hat{\mathbf{x}} = [x, y, \theta, v, \omega]^\top$ at 10\,Hz.

To handle network variability, the cloud continuously estimates each node’s communication delay online as
\begin{equation}
    \Delta t_n \sim \mathcal{N}(\mu_n,\, \sigma_n^2),
    \label{eq:latency_model}
\end{equation}
where $\mu_n$ and $\sigma_n$ are computed from a sliding window of recent message timestamps. 

Incoming data is classified as either \emph{normal} (arriving within the adaptive fusion window) or \emph{late}. Rather than discarding delayed messages, the cloud compensates for latency by propagating object states to the current fusion time using EKF-based motion prediction before integration. Data with excessive delay is incorporated with reduced confidence or excluded for that cycle, ensuring robustness under temporary congestion while maintaining global perception continuity.


\subsection{Infrastructure Sensing vs Onboard Perception}
\begin{figure}
    \centering
    \includegraphics[width=0.4 \textwidth]{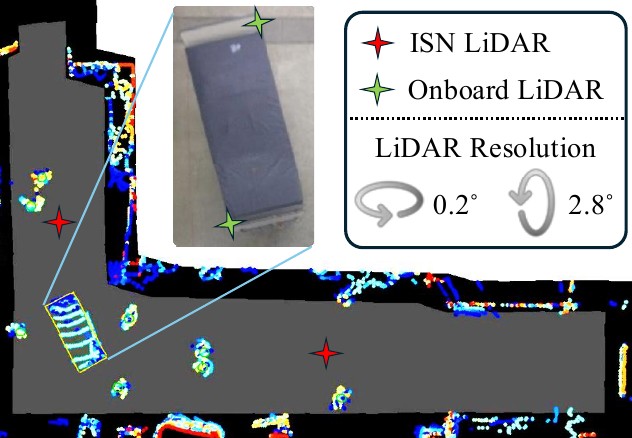}
    \caption{Placements of ISN and Onboard LiDARs}
    \label{Fig: onboardvsisn}
\end{figure}

A key advantage of the CAM system is the elevated and fixed placement of its ISNs, which provide a clear and stable view of the environment even in the presence of large vehicles, narrow corridors, or crowded scenes.
To quantify this advantage, we compare the ISN sensing against an onboard sensing configuration using the same sensor type and detection thresholds as shown in Fig.~\ref{Fig: onboardvsisn}.
The evaluation is conducted on a dataset collected in a hallway environment where an autonomous robot navigates among eight pedestrians, a scenario that frequently leads to occlusions, as pedestrians overlap along the line of sight of onboard sensors mounted at corner positions.



\begin{figure}[t]
    \centering
    \begin{subfigure}[t]{0.45\textwidth}
        \centering
        \includegraphics[width=\textwidth]{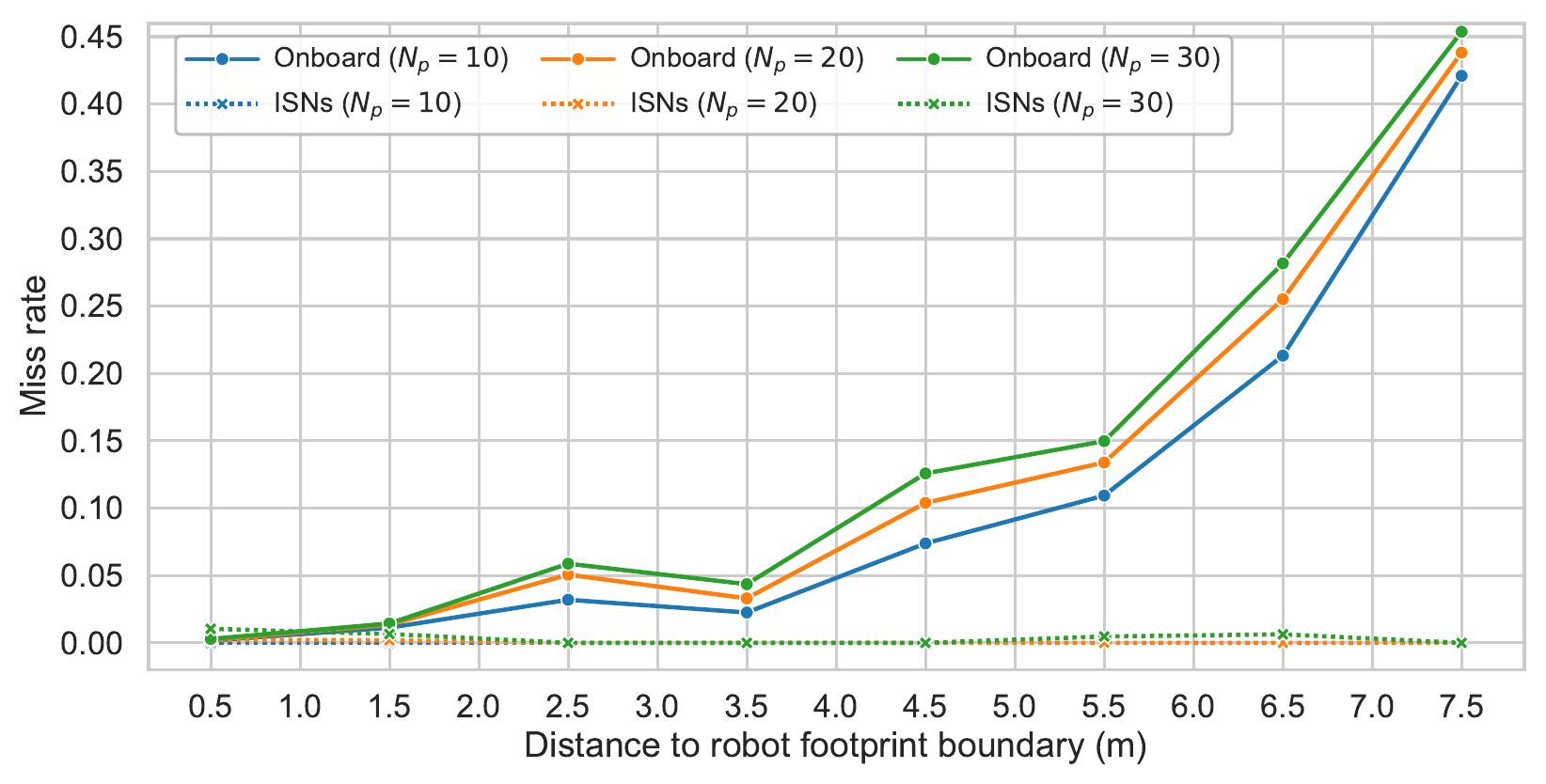}
        \caption{Miss rate vs distance under different $N_p$.}
        \label{fig:missrate_sub}
    \end{subfigure}
    \hfill
    \begin{subfigure}[t]{0.4\textwidth}
        \centering
        \includegraphics[width=\textwidth]{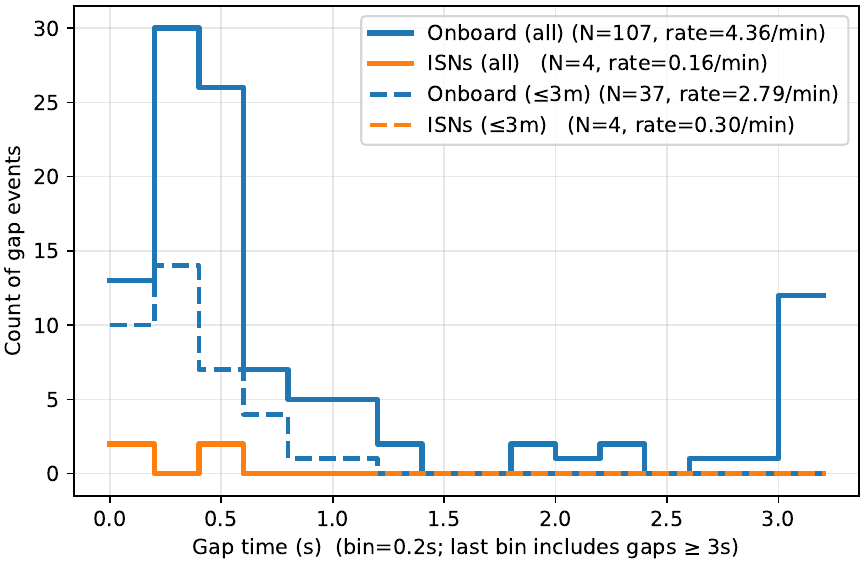}
        \caption{Gap duration distribution at $N_p=20$.}
        \label{fig:gap_sub}
    \end{subfigure}
    \caption{Perception reliability comparison between Onboard and ISN configurations. 
    (a) Spatial miss rate as vs distance to robot footprint boundary.
    (b) Temporal gap distribution, with dashed lines indicating the $\leq 3\,\mathrm{m}$ critical zone.}
    \label{fig:reliability_combined}
\end{figure}

\paragraph{Evaluation metrics.}
Two complementary metrics capture different dimensions of detection failure.
The \textbf{miss rate} measures the fraction of ground-truth person instances that remain undetected for a given distance bin and point-count threshold $N_p$, where $N_p$ is the minimum number of LiDAR returns required for a cluster to make a valid detection.
\textbf{Temporal consistency} is assessed via \textbf{gap events}: along each ground-truth track, a gap is a maximal continuous sequence of missed detections bounded on both sides by successful detections, summarized by its duration in seconds.
Beyond the total gap count~($N$), we report a normalized \textbf{gap rate} in events per minute of track time, enabling fair comparison across evaluation subsets with different temporal extents.

\paragraph{Detection reliability vs.\ distance.}
Fig.~\ref{fig:reliability_combined} (a) shows how miss rate evolves with increasing distance from the target to the robot footprint boundary, evaluated over three thresholds ($N_p \in \{10, 20, 30\}$). 
Across all thresholds, the ISN configuration maintains near-zero miss rate over the full evaluated range, indicating that elevated sensing largely suppresses the distance-dependent degradation caused by occlusion. 
In contrast, the onboard system exhibits a monotonically increasing miss rate that exceeds $0.40$ at $7.5\,\mathrm{m}$ for all three thresholds.
This degradation is primarily caused by line-of-sight occlusion: as distance increases, pedestrians are more likely to overlap along the viewing direction of the corner-mounted onboard sensors, causing farther individuals to be partially or fully blocked by those closer to the robot.
The elevated placement of ISN sensors largely mitigates this effect, maintaining clearer visibility and preventing the distance-dependent failures.

\paragraph{Temporal consistency and gap analysis.}
Miss rate alone cannot distinguish isolated single-frame failures from sustained tracking dropouts, which carry very different implications for downstream planning.
Fig.~\ref{fig:reliability_combined} (b) presents the gap duration distribution at $N_p=20$, including a focused analysis of the \textbf{critical zone} ($\leq 3\,\mathrm{m}$ from person center to robot boundary) where perception failures have great impacts on safe navigation.
The difference is substantial: the onboard system accumulates $107$ gap events overall ($4.36\,\mathrm{events/min}$), compared to just $4$ for the ISN configuration ($0.16\,\mathrm{events/min}$).
Within the critical zone, the onboard count stands at $37$ ($2.79\,\mathrm{events/min}$) while the ISN count remains at $4$ ($0.30\,\mathrm{events/min}$).
Although most onboard gaps are relatively short (with a mode between $0.2$ and $0.6\,\mathrm{s}$), their frequency is significantly higher than that of the ISN system.
In contrast, ISN gaps are rare and brief-concentrated below $0.4\,\mathrm{s}$ consistent with its near-zero miss rates and improved temporal consistency.

\paragraph{Pose estimation accuracy.}
Beyond occlusion robustness, the accuracy of ISN-generated position and orientation estimates is critical for trajectory planning.
The cooperative camera–LiDAR fusion pipeline evaluated in the prior work~\cite{ning2025sapcopesocialawareplanningusing} achieved a localization mean absolute error (MAE) of approximately $8\,\mathrm{cm}$ and a yaw MAE of approximately $7^{\circ}$ across multiple ISNs, outperforming both single-node and learning-based baselines.
A delay-aware fusion strategy further maintained these accuracy levels under communication latencies up to $200\,\mathrm{ms}$, confirming that ISN-based pose estimates are sufficiently reliable to serve as the perception backbone for real-time motion planning in crowded environments.

\subsection{5G Communication and Latency}
\label{sec:latency_analysis}

Achieving real-time cooperative perception over a cellular network requires not only low average latency but predictable, consistent behaviour under continuous, large-scale operation.
To characterize both properties, end-to-end latency was measured across $839{,}951$ messages from 14 outdoor sensor nodes over approximately 100 minutes of continuous field deployment.
The end-to-end latency decomposes into two components: the \emph{local processing latency} (sensor acquisition to perception output at the edge unit) and the \emph{5G communication latency} (perception message transmission from node to cloud).

\begin{figure}[t]
    \centering
    \begin{subfigure}[t]{0.4\textwidth}
        \centering
        \includegraphics[width=\textwidth]{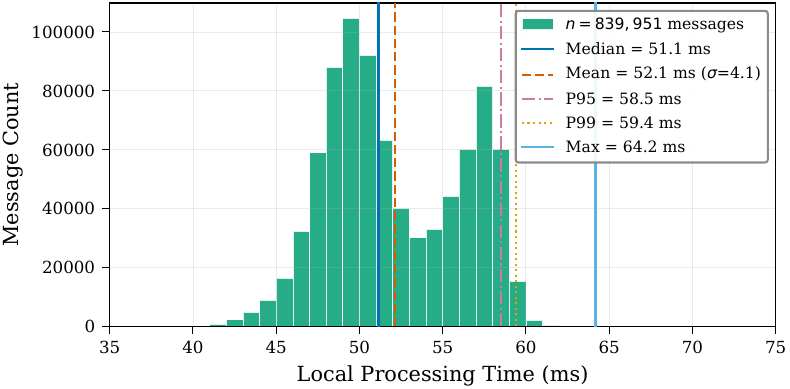}
        \caption{Local processing latency.}
        \label{Fig: latency_for_local_perception}
    \end{subfigure}
    \hfill
    \begin{subfigure}[t]{0.4\textwidth}
        \centering
        \includegraphics[width=\textwidth]{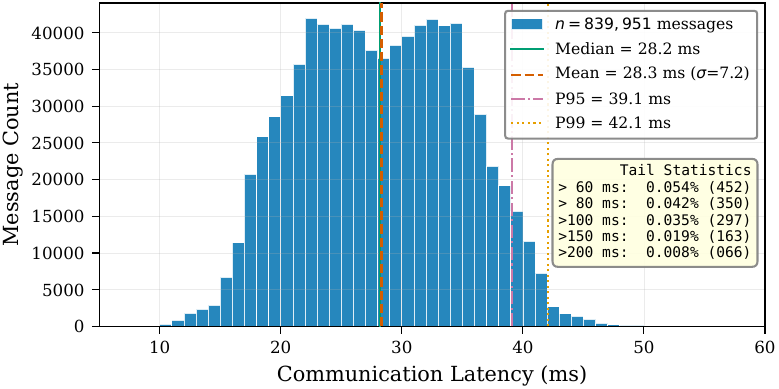}
        \caption{5G communication latency. 99.95\% $\leq$ 60\,ms.}
        \label{Fig: latency_for_5g_network}
    \end{subfigure}

    \vspace{0.5em}

    \begin{subfigure}[t]{0.38\textwidth}
        \centering
        \includegraphics[width=\textwidth]{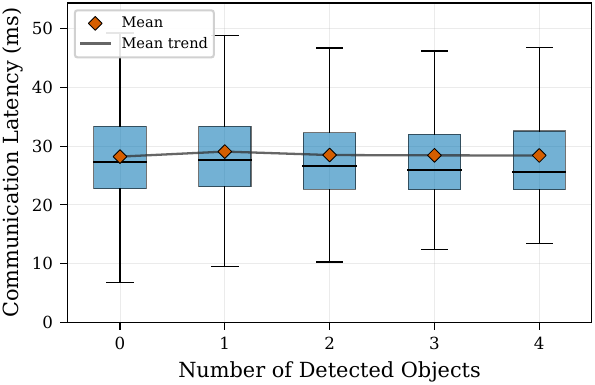}
        \caption{5G latency vs detected objects per message. }
        \label{Fig: latency_for_5g_network_objects}
    \end{subfigure}

    \caption{Latency characterization of the 5G CAM system. 
    (a) Local perception processing latency across all nodes. 
    (b) 5G communication latency distribution. 
    (c) Communication latency versus number of detected objects per message.}
    \label{fig:latency_combined}
\end{figure}

\paragraph{Local processing latency.}
Fig.~\ref{fig:latency_combined} (a) shows the distribution of local processing latency across all 14 nodes.
The pipeline encompassing image and point cloud preprocessing, YOLOv11s-OBB \cite{yolov11_ultralytics} inference with TensorRT acceleration, and message serialization exhibits a tightly concentrated distribution with a mean of $52.1\,\mathrm{ms}$ ($\sigma = 4.1\,\mathrm{ms}$) and a median of $51.1\,\mathrm{ms}$.
The 99th percentiles reach only $59.4\,\mathrm{ms}$, with an observed maximum of $64.2\,\mathrm{ms}$.
The narrow spread across nodes and operating conditions confirms that the distributed edge hardware delivers stable, consistent real-time throughput during the deployment.

\paragraph{5G communication latency.}
Fig.~\ref{fig:latency_combined} (b) presents the 5G communication latency distribution.
The mean is $28.3\,\mathrm{ms}$ ($\sigma = 7.2\,\mathrm{ms}$) with a median of $28.2\,\mathrm{ms}$; the 95th and 99th percentiles are $39.1\,\mathrm{ms}$ and $42.1\,\mathrm{ms}$.
Crucially, only $0.054\%$ of messages exceeded $60\,\mathrm{ms}$, and fewer than $0.01\%$ exceeded $200\,\mathrm{ms}$, demonstrating exceptional link reliability throughout the full duration of the deployment.

A practical concern in cooperative perception is whether increased message payload, caused by detecting more objects, introduces additional communication delay.
Fig.~\ref{fig:latency_combined} (c) evaluates this effect by grouping latency according to the number of detected objects per message.
Each additional object appends approximately $1{,}151$ bytes (bounding box, classification, and tracking metadata) to a $44$-byte base header.
Despite this increase, the mean latency remains effectively constant.
This suggests that, within the payload range of this deployment, 5G latency is dominated by fixed network-layer overhead, such as radio scheduling and core network traversal, rather than payload transmission time, supporting scalability as scene complexity grows.

For reference, local Wi-Fi achieves a communication latency of $2.3\,\mathrm{ms}$ ($\sigma = 0.3\,\mathrm{ms}$), substantially lower due to the absence of cellular core network traversal.
This difference is expected and highlights the trade-off: the CAM system incurs a modest latency increase in exchange for wide-area coverage and outdoor deployment flexibility that Wi-Fi cannot provide.

\paragraph{End-to-end performance.}
Summing both components, the total end-to-end latency from sensor data acquisition to cloud availability is $80.4\,\mathrm{ms}$ via 5G ($52.1 + 28.3\,\mathrm{ms}$), or $54.4\,\mathrm{ms}$ via Wi-Fi ($52.1 + 2.3\,\mathrm{ms}$).
For cooperative perception applications requiring sub-$100\,\mathrm{ms}$ response to support real-time autonomous decision-making, the 5G-based ISN system meets this requirement for approximately $99\%$ of messages, a strong empirical confirmation of the system's readiness for real-world large-scale deployment.




\subsection{The Accuracy--Latency Balance: Fusion Strategy as a Design Choice}

A key architectural decision in cooperative perception is how to fuse multi-node sensor data. At one extreme, \emph{early fusion} aggregates raw multi-view sensor data before detection, enabling feature-level integration across nodes. At the other extreme, \emph{late fusion} performs detection independently at each node and transmits only compact object-level outputs for global association. In practice, fusion lies on a spectrum: intermediate strategies, such as sharing compressed feature maps or lightweight BEV encodings, can trade communication cost against detection performance.

Early fusion represents the accuracy upper bound of the system, as multi-view feature aggregation can recover partially occluded or sparsely observed objects before the detection stage. Late fusion, by contrast, significantly reduces bandwidth consumption and communication overhead, making it attractive for large-scale deployments with constrained network resources. The CAM platform provides a real-world testbed to quantify this accuracy–latency trade-off across object classes and weather conditions.

Using the CoInfra dataset~\cite{ning2025coinfra}, we compare an \emph{early-fusion} approach (GlobalLidarCamOBB), which aggregates LiDAR and camera data from all nodes into a global multi-channel BEV representation prior to detection, with a \emph{late-fusion} approach (LocalLidarCamOBB), which performs per-node LiDAR–camera BEV detection and merges oriented bounding boxes via NMS. Both methods use YOLOv11s-OBB \cite{yolov11_ultralytics} as the detection backbone. Performance is reported as AP@0.5 for \textbf{Vehicle} (car, SUV, pickup, van, bus) and \textbf{VRU} (pedestrian, cyclist, scooter).

\begin{table}[t]
\centering
\caption{AP@0.5 versus distance to the closest node for Vehicle and VRU.}
\label{tab:coinfra_distance_ap50}
\setlength{\tabcolsep}{6pt}
\begin{tabular}{llcccc}
\toprule
\textbf{Method} & \textbf{Class} & \textbf{0--15m} & \textbf{15--25m} & \textbf{25--35m} & \textbf{35--50m} \\
\midrule
\multirow{2}{*}{Global}
& Vehicle & 1.000 & 0.998 & 0.998 & 0.992 \\
& VRU     & 0.660 & 0.687 & 0.611 & 0.548 \\
\midrule
\multirow{2}{*}{Local}
& Vehicle & 0.994 & 0.920 & 0.817 & 0.753 \\
& VRU     & 0.540 & 0.567 & 0.390 & 0.261 \\
\bottomrule
\end{tabular}
\end{table}

\begin{table}[t]
\centering
\caption{Weather robustness (AP@0.5) for Vehicle and VRU. `All' aggregates all weather conditions.}
\label{tab:coinfra_weather_ap50}
\setlength{\tabcolsep}{7pt}
\begin{tabular}{llccc}
\toprule
\textbf{Method} & \textbf{Class} & \textbf{Sunny} & \textbf{Heavy Snow} & \textbf{All} \\
\midrule
\multirow{2}{*}{Global} 
& Vehicle & 0.996 & 0.995 & 0.995 \\
& VRU     & 0.654 & 0.585 & 0.618 \\
\midrule
\multirow{2}{*}{Local} 
& Vehicle & 0.918 & 0.907 & 0.908 \\
& VRU     & 0.423 & 0.326 & 0.364 \\
\bottomrule
\end{tabular}
\end{table}

\paragraph{Distance sensitivity.}
Table~\ref{tab:coinfra_distance_ap50} reports AP@0.5 as a function of distance to the closest sensor node.
For \emph{Vehicle} detection, early fusion maintains near-saturated performance (AP@0.5~$\approx$~1.0) across all distance bins, while late fusion starts around 0.92 at the closest range and decreases gradually.
The gap remains moderate for large, rigid objects, suggesting that late fusion is viable when vehicles are the primary concern and bandwidth is limited.
For \emph{VRU} detection, however, the divergence is substantial. The performance gap widens sharply with distance: early fusion maintains moderate accuracy, whereas late fusion drops rapidly beyond mid-range distances. This behaviour reflects a key mechanism: small and partially occluded objects that are weakly observed at any single node can be recovered through multi-view feature aggregation in early fusion, but are often irretrievably lost once a per-node detector fails in late fusion.

\paragraph{Weather robustness.}
Table~\ref{tab:coinfra_weather_ap50} evaluates performance under adverse weather. For Vehicle detection, early fusion achieves near-perfect AP@0.5 in both Sunny (0.996) and Heavy Snow (0.995) conditions, with minimal degradation; late fusion remains competitive (0.918 Sunny, 0.907 Heavy Snow). 
For VRUs, the contrast is more pronounced. Early fusion achieves 0.654 in Sunny and 0.585 in Heavy Snow conditions, while late fusion achieves only 0.423 and 0.326, respectively. The amplified gap under heavy snow, where LiDAR attenuation, reduced visibility, and partial occlusion disproportionately affect small objects, highlights the robustness advantage of infrastructure-level feature aggregation.

\paragraph{Design guidance.}
Aggregated across conditions, early fusion outperforms late fusion by nearly $1.7\times$ for VRUs (0.618 vs.\ 0.364 AP@0.5), while the margin for Vehicles is smaller but consistent (0.995 vs.\ 0.908). These results provide practical design guidance: when detection of vulnerable or small objects is safety-critical, early fusion is strongly preferred. When bandwidth or computational constraints dominate and large vehicles are the primary target, late fusion remains a deployable alternative.
The pronounced gap between the two extremes, especially for VRUs under adverse weather, motivates exploration of intermediate fusion strategies that balance feature sharing and communication cost. The CAM platform makes this accuracy–latency frontier explicit, enabling principled fusion strategy selection based on application requirements.



\section{Case Study: Outdoor application} \label{Chap: Outdoor application}



The proposed CAM framework is validated in a real urban deployment: a four-way roundabout on a local city road with 14 ISNs mounted on roadside light poles and connected to a cloud over a commercial 5G, Fig. \ref{Fig: framework} (c) demonstrates one of the outdoor ISNs. The deployment is spatially structured: six ISNs cover the straight approaches, and eight surround the roundabout to maximize visibility of merging and yielding behaviours as shown in Fig. \ref{Fig: framework} (a). This site is a practical stress test for cooperative perception, with frequent occlusions from buildings and trees, dense traffic participant interactions, and rapidly changing right-of-way.



\subsection{Road Hazard and Critical Scenario Warning}
In roundabout traffic, many critical events are interaction-driven \cite{shu2023human}: a maneuver that appears safe for one participant can become hazardous when another enters the conflict region. CAM allows such interactions to be detected at the cloud level and broadcast as warnings to connected endpoints.

\begin{figure}[t]
    \centering
    \begin{subfigure}[b]{0.24\textwidth}
        \centering
        \includegraphics[width=\textwidth]{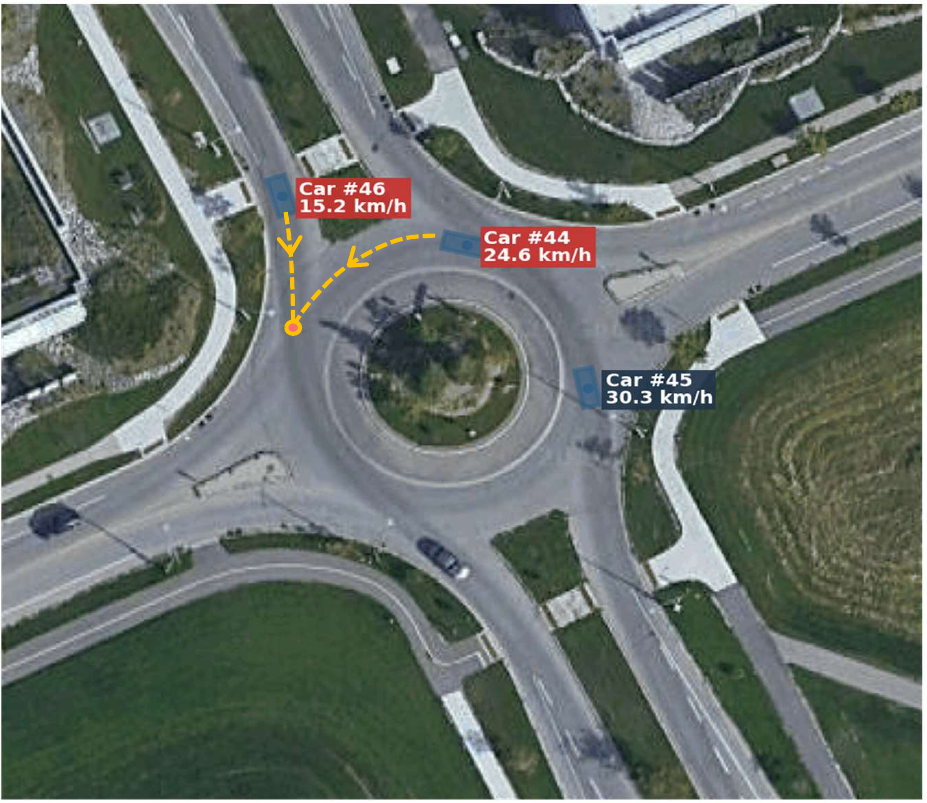}
        \caption{Vehicle-Vehicle Interaction.}
        \label{fig:road_hazard_image1}
    \end{subfigure}
    \hfill
    \begin{subfigure}[b]{0.24\textwidth}
        \centering
        \includegraphics[width=\textwidth]{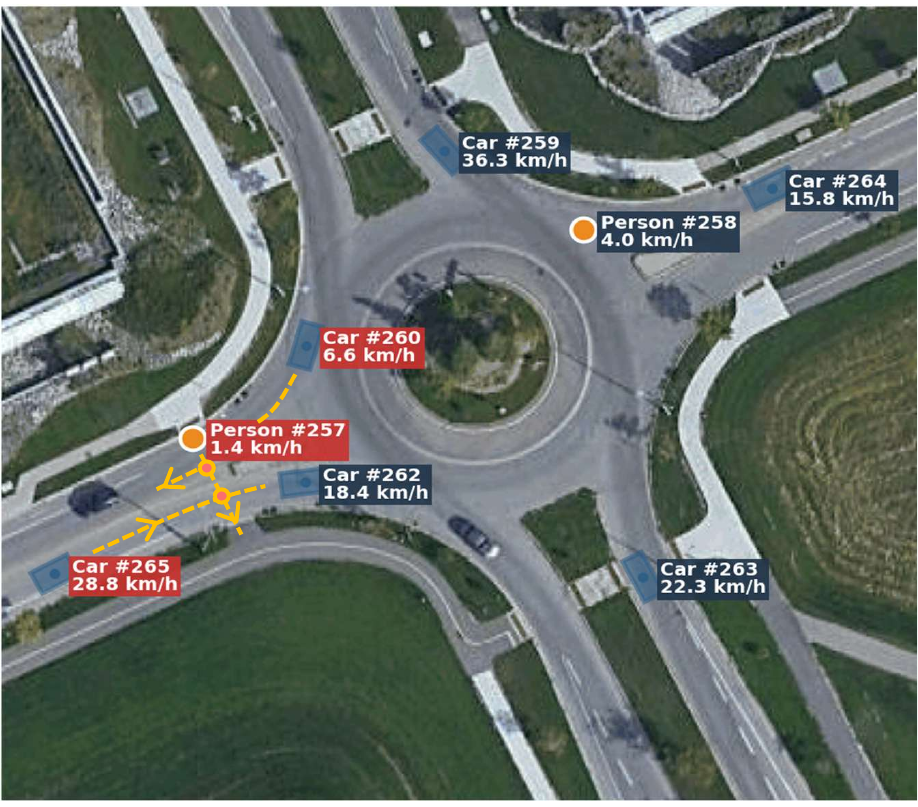}
        \caption{Vehicle-Pedestrian Interaction.}
        \label{fig:road_hazard_image2}
    \end{subfigure}
    \caption{Examples of safety warning scenarios supported by the outdoor CAM deployment.}
    \label{fig:road_hazard_overall_label}
\end{figure}

In this case, a Signal Temporal Logic (STL) and game-theoretic–based module is implemented to issue alerts for scenarios such as potential conflicts, blocked lanes, and rule-violation risks. One representative example is a vehicle entering the roundabout while a pedestrian is attempting to cross at an approach, creating a potential collision scenario. CAM can generate a timely warning that is deliverable either to a connected vehicle or to a 5G-enabled mobile application. Fig.~\ref{fig:road_hazard_overall_label} provides representative examples for vehicle–vehicle and vehicle–pedestrian interactions. These warnings complement onboard autonomy by explicitly targeting the failure modes that arise from occlusions and limited line of sight in complex urban geometry.

\subsection{Connected Autonomous Driving}

Beyond warning services, the same cooperative perception capability can support connected autonomous driving. The main advantage of CAM is \emph{visibility beyond the vehicle}: infrastructure sensors observe the scene from elevated viewpoints and can detect traffic participants that are partially or fully occluded from the ego vehicle's onboard sensors. Through the 5G connection, this information can be delivered to connected vehicles with sufficiently low latency to support driving decisions.

\begin{figure}[tb]
    \centering
    \includegraphics[width=0.49\textwidth]{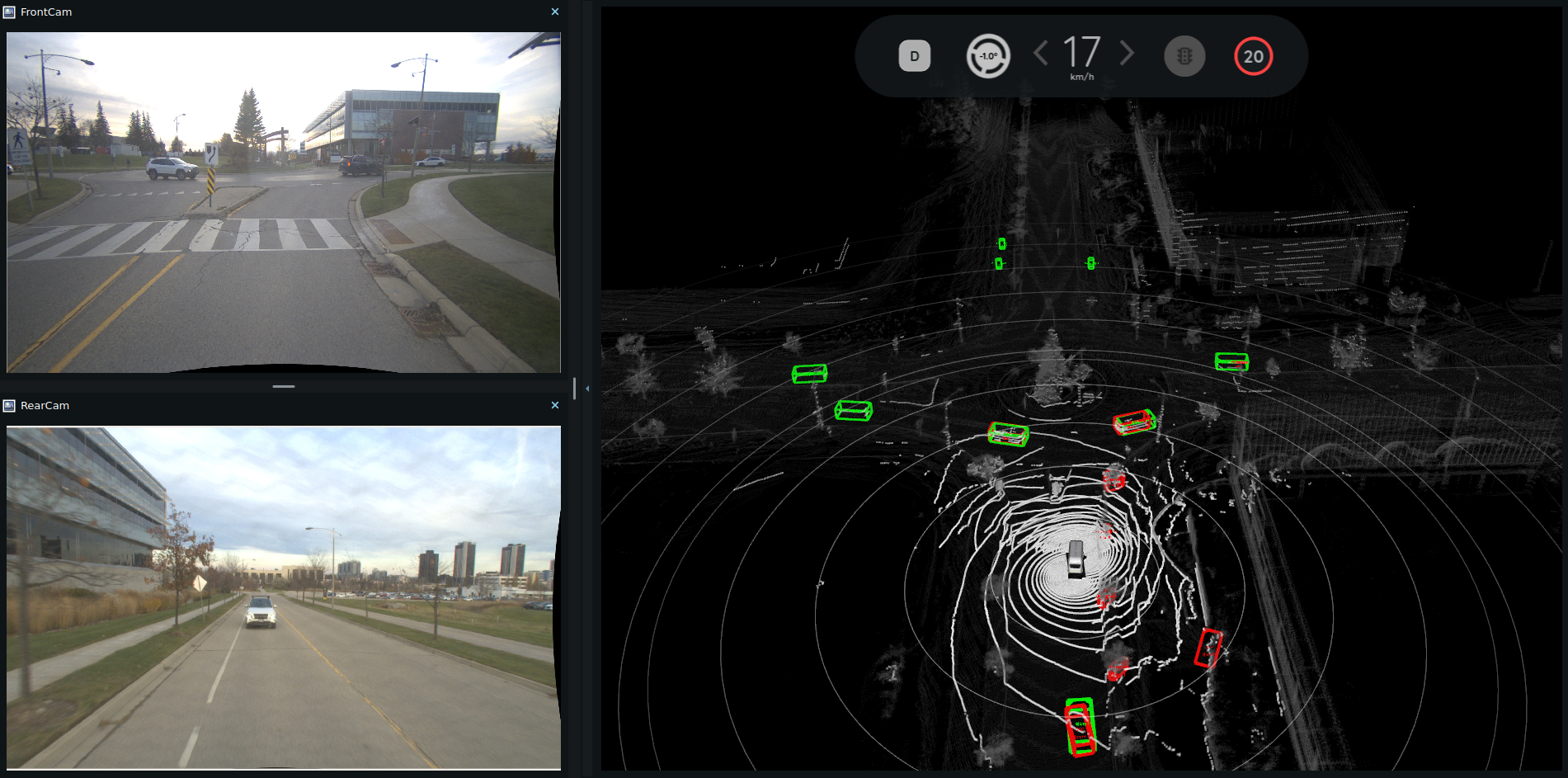}
    \caption{Cloud-enabled cooperative perception for autonomous driving. 
    Left: front and rear camera views from the vehicle. 
    Right: fused LiDAR perception, where gray points denote the HD map, 
    white points are vehicle LiDAR returns, red boxes are onboard detections, 
    and green boxes are detections from cloud-connected infrastructure sensors.}
    \label{Fig: coinfrabusautoware}
\end{figure}

To demonstrate this capability, the cloud-generated cooperative perception output is integrated with an autonomous shuttle bus through Autoware~\cite{autoware} using a LiDAR--camera--HD map fusion pipeline \cite{ning2024efficient}. Fig.~\ref{Fig: coinfrabusautoware} shows an example of this vehicle-side interface. Onboard detections are shown in red, while cloud-connected infrastructure detections are shown in green. The combined result provides a more complete understanding of nearby traffic participants than onboard sensing alone, especially in occlusion-prone areas around the roundabout.

This shared scene understanding also provides a foundation for interaction-aware decision making\cite{shu2024game,shu2025decision}. The global traffic state can be interpreted with respect to the roundabout topology and traffic rules, allowing interaction risks to be evaluated in a structured way. In our system, STL-based reasoning is used to quantify rule compliance and interaction safety margins among traffic participants, which can then inform higher-level planning and control modules.

As a result, a connected autonomous vehicle can react not only to immediate collision risks, but also to emerging interaction hazards such as unsafe entry timing, failure to yield, or pedestrian conflicts. By combining cooperative perception with rule-aware interaction reasoning, the CAM framework extends connected driving beyond perception enhancement alone and moves toward safer and more human-compatible behavior in complex urban traffic.

\section{Case Study: Indoor application} \label{Chap: Indoor application}


To assess the feasibility of the proposed CAM framework for the indoor environment, the proposed system was implemented and evaluated in our lab to simulate the hospital environment. As shown in Fig. \ref{Fig: framework} (b), a total of four sensor nodes are installed across two separate areas for different experimental objectives. Specifically, Nodes i, ii, and iii are deployed along a hallway to evaluate autonomous driving functionalities, whereas Node iv is positioned to assess human-robot interaction ability. Fig. \ref{Fig: framework} (d) shows each individual indoor ISNs. 


An autonomous robot that was retrofitted from a medical bed, shown in Fig.~\ref{Fig: framework} (f), is utilized for both testing scenarios. The robot is equipped with two driving modules: one mounted at the front and the other at the rear. Each module contains a centrally located motorized wheel, flanked by two caster wheels on either side for stability. The motorized wheels are capable of generating independent driving forces and steering motions, enabling omnidirectional movement of the robot.

\subsection{Autonomous Motion Tracking}
\begin{figure}
    \centering
    \includegraphics[width=0.49 \textwidth]{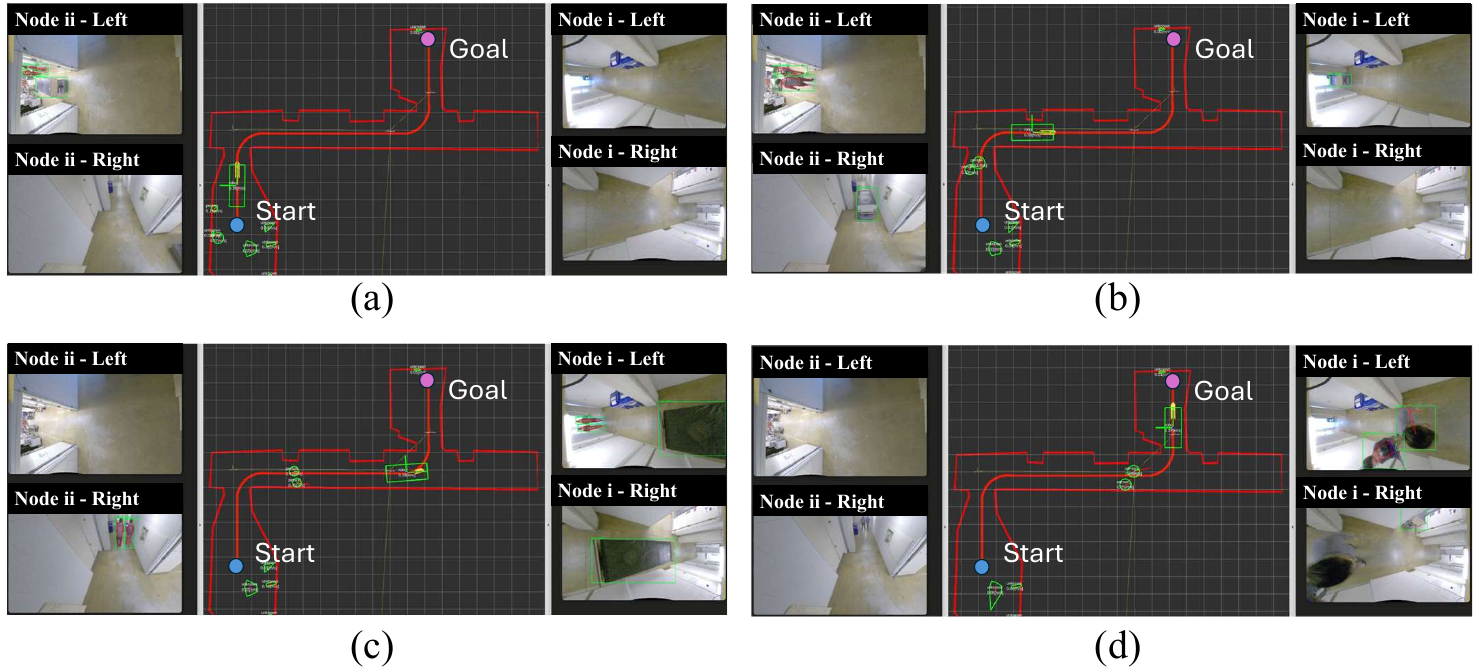}
    \caption{Autonomous Motion Tracking Results. (a) t = 2 sec. (b) t = 21 sec. (c) t = 39 sec. (d) t = 52 sec.}
    \label{Fig: Indoor_tracking_results}
\end{figure}
For the autonomous driving evaluation, the robot was controlled by a model predictive control (MPC) controller \cite{yang2024intelligent}, starting from the defined starting position to reach the goal position. As the robot moved through the hallway, real-time position estimates were generated through the cooperative sensing framework, enabling accurate tracking as the robot transitioned between the coverage zones of adjacent nodes. 

Fig.~\ref{Fig: Indoor_tracking_results} presents images captured by ISN i, ii and the robot pose at selected time instances during the experiment. As shown, the proposed perception framework is able to detect and identify the robot accurately using trained neural networks. Based on these detections, the corresponding positions of the robot are then estimated by fusing the camera outputs with LiDAR point cloud data. In this particular example, the system focuses on tracking the medical-bed shape robot; however, the same framework can be extended to recognize and track other types of robots. 

\subsection{Interactive-aware Motion Planning}
A robust interactive-aware motion planner is implemented for this scenario. At each time step, the positions and poses of the robot and nearby pedestrians are estimated by Node iv and transmitted to the Cloud Computation Unit. An MPC planner simultaneously predicts pedestrian intentions and computes appropriate control commands for the robot.
\begin{figure}
    \centering
    \includegraphics[width=0.49 \textwidth]{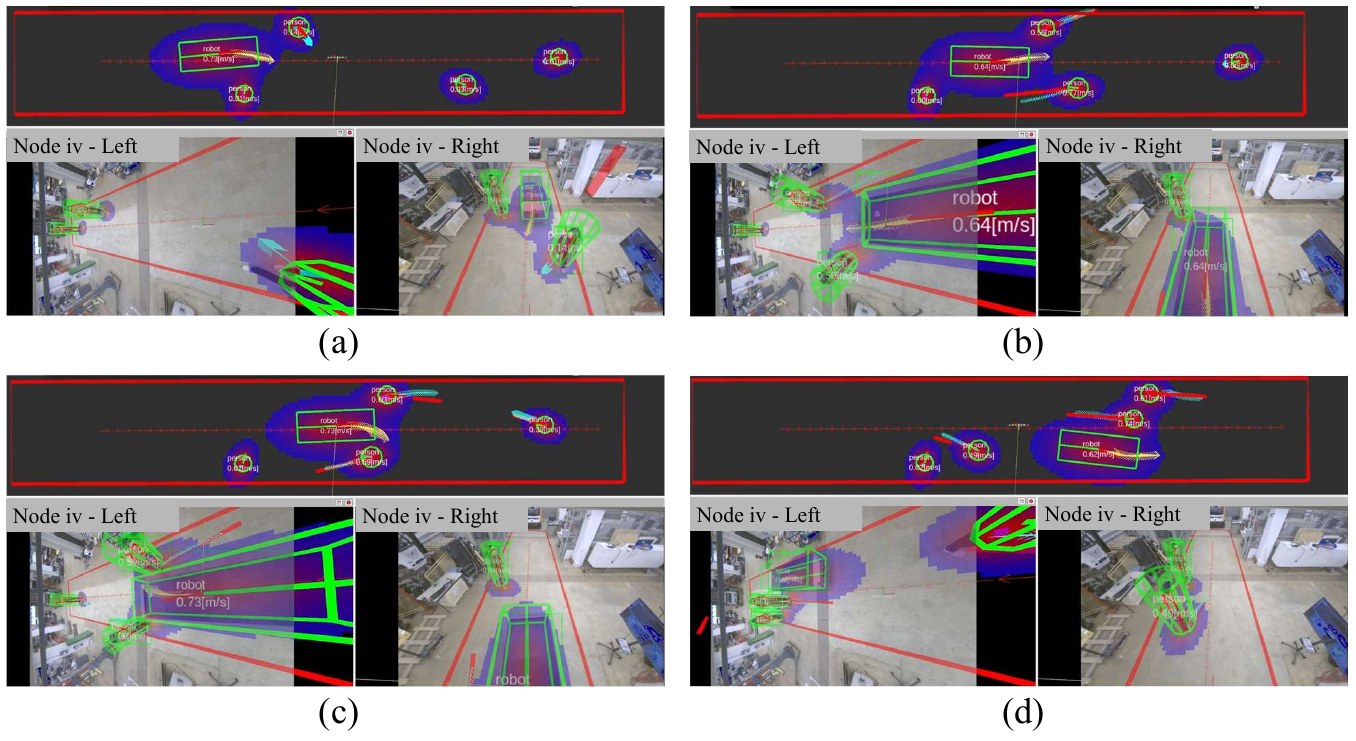}
    \caption{Interactive-aware Motion Planning Results. (a) t = 5 sec. (b) t = 8 sec. (c) t = 12 sec. (d) t = 15 sec.}
    \label{Fig: Indoor_results}
\end{figure} 

Fig. \ref{Fig: Indoor_results} presents the interactive-aware motion planning results from a single experiment at four different time instances (a)–(d), illustrating the temporal evolution of the robot’s behaviour in a dynamic, human-populated indoor environment. In each sub-figure, the top panel shows a global top-view occupancy map. The robot is depicted as a green rectangle, while the surrounding pedestrians are represented by circular boundaries. The red dashed line denotes the predefined reference path.

The yellow and cyan arrow sequences indicate the predicted intentions of the robot and pedestrians, respectively, over the prediction horizon. For comparison, the solid red arrow associated with each pedestrian represents the predicted intention generated by a simple linear model that assumes constant walking speed and heading. The shaded regions, ranging from deep blue to red, correspond to the interaction-aware collision cost field generated by all agents, including both the robot and pedestrians. Warmer colors indicate higher interaction intensity and potential collision risk, effectively capturing the dynamic coupling among agents. The lower panels display the real camera images along with the projected perception results from ISN iv. 

From (a) to (d), the robot successfully avoids all pedestrians while maintaining close tracking of the predefined reference path. Notably, the interaction-aware predictions (cyan arrows) exhibit more realistic and adaptive behaviour compared to the simple linear predictor, particularly in scenarios involving mutual avoidance or trajectory adjustments. These results demonstrate that the proposed framework enables safe, smooth, and socially compliant navigation through continuous intention estimation and interaction-aware motion planning.

\section{Conclusion} \label{Chap: Conclusion}

This article introduced the Cloud-based Autonomous Mobility (CAM) framework, a generalized infrastructure-based cloud architecture designed to support autonomous systems across both outdoor and indoor environments. By combining high-mounted multi-modal sensing, edge computing, and low-latency wireless communication, the proposed Intelligent Sensor Nodes (ISNs) convert infrastructure from a passive physical asset into an active perception layer that augments onboard autonomy.

The real-world deployments demonstrate the effectiveness and practical impact of the framework. In the outdoor roundabout deployment with 14 ISNs, CAM enabled cloud-level global perception fusion, providing visibility beyond individual vehicles. The fused scene representation supported cooperative perception for connected autonomous vehicles and facilitated proactive hazard warnings in interaction-driven scenarios such as vehicle–vehicle and vehicle–pedestrian conflicts. These results highlight the framework’s ability to mitigate occlusion and limited line-of-sight issues inherent in complex urban geometries.

In the indoor hospital-simulated environment, the system supported both autonomous motion tracking and interactive-aware navigation. Infrastructure-assisted perception enabled accurate robot localization and pedestrian detection across multiple node coverage regions and facilitated socially compliant motion planning in dynamic human-populated corridors. The interactive-aware MPC demonstrated adaptive intention prediction and smooth collision avoidance while maintaining reference path tracking, outperforming simple linear prediction models in realistic human–robot interaction scenarios.

Together, these case studies validate the CAM framework as a scalable and unified solution for distributed perception across different environments. Beyond the specific hardware and software implementations presented here, the broader contribution of this work lies in advancing a shift toward infrastructure–robot collaboration. Rather than relying solely on increasingly complex onboard sensing stacks, future autonomous mobility systems can benefit from shared environmental intelligence that is scalable, cooperative, and resilient.

\section*{Acknowledgment}
The authors gratefully acknowledge the financial support of the Natural Sciences and Engineering Research Council of Canada (NSERC)  and MITACS, as well as the financial and technical support provided by Rogers Communications Inc. Canada.


 
%
\bibliographystyle{IEEEtran}
\bibliography{Reference.bib}

\vfill

\end{document}